\documentclass[conference,letterpaper]{IEEEtran}

\usepackage{amsmath,amssymb,amsfonts}
\usepackage{graphicx}
\usepackage{textcomp}
\usepackage{xcolor}
\usepackage{hyperref}
\usepackage{booktabs}
\usepackage{algorithm}
\usepackage{algorithmic}
\usepackage{multirow}
\usepackage{subcaption}

\newcommand{\zt}{z_t}
\newcommand{\ot}{o_t}
\newcommand{\at}{a_t}
\newcommand{\hist}{h_t}
\newcommand{\ct}{c_t}
\newcommand{\methodname}{\textsc{RecoverFormer}}

\begin{document}

\title{RecoverFormer: End-to-End Contact-Aware Recovery\\ for Humanoid Robots}

\author{
\IEEEauthorblockN{Zihui Liu}
\IEEEauthorblockA{
\textit{Stanford University} \\
zl90@alumni.stanford.edu}
}

\maketitle

% ============================================================
\begin{abstract}
Humanoid robots operating in unstructured environments must recover from unexpected disturbances---a capability that remains challenging for end-to-end control policies.
We present \methodname{}, a fully end-to-end humanoid recovery policy that learns \emph{when} and \emph{how} to switch among recovery behaviors---including compensatory stepping, hand-environment contact, and center-of-mass reshaping---while maintaining robust performance under model mismatch.
The architecture combines a causal transformer over a 50-step observation history with two novel heads: a \emph{latent recovery mode} that enables smooth transitions among distinct recovery strategies, and a \emph{contact affordance head} that predicts which environmental surfaces (walls, railings, table edges) are beneficial for stabilization.
We evaluate \methodname{} on the Unitree G1 humanoid in MuJoCo.
Trained only on open floor, \methodname{} transfers \emph{zero-shot} to walled environments, achieving $\mathbf{100\%}$ recovery success across 100--300\,N pushes and across wall distances from $0.25$--$1.4$\,m.
Under zero-shot dynamics mismatch, \methodname{} reaches $75.5\%$ at $+25\%$ mass, $89\%$ under $30$\,ms latency, $91.5\%$ at low friction, and $99\%$ under compound friction$+$latency$+$mass perturbation.
The learned latent modes specialize across force regimes without mode-level supervision, validated by t-SNE analysis of 300 episodes.
Taken together, these results show that a single end-to-end policy can deliver multi-modal, contact-aware humanoid recovery that generalizes across perturbation magnitude, contact geometry, and dynamics shift.
\end{abstract}

\begin{IEEEkeywords}
Humanoid robots, fall recovery, contact-aware control, end-to-end learning, reinforcement learning
\end{IEEEkeywords}

\begin{figure}[t]
    \centering
    \includegraphics[width=0.55\columnwidth]{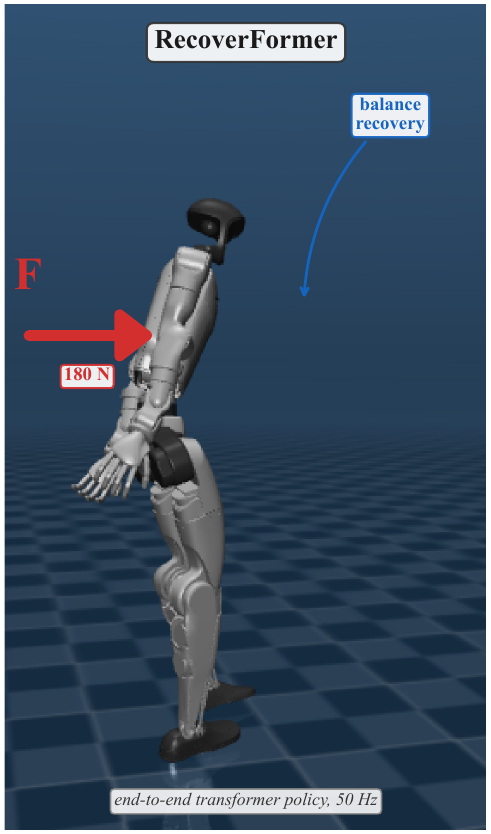}
    \caption{\textbf{\methodname{} maintains balance through a lateral push on the Unitree G1.}
      Mid-recovery snapshot $160$\,ms after a $180$\,N torso impulse.
      The transformer policy converts a $50$-step history of proprioception and contact-region distances into $29$-DoF joint targets at $50$\,Hz.}
    \label{fig:teaser}
\end{figure}

% ============================================================
\section{Introduction}
\label{sec:intro}

Humanoid robots are increasingly deployed in unstructured human environments where unexpected perturbations---collisions, slippery surfaces, payload shifts---can destabilize the robot and cause falls.
Unlike wheeled or quadruped platforms, humanoid robots have a narrow support polygon and high center of mass, making balance recovery a safety-critical capability~\cite{wang2017realtime}.

\textbf{The recovery problem.}
When a humanoid is pushed beyond its balance limits, effective recovery requires choosing among qualitatively different strategies: taking a compensatory step, bracing against a nearby surface, or rapidly lowering the center of mass.
Humans perform this selection seamlessly, exploiting environmental affordances---a wall to lean on, a railing to grasp---in a fraction of a second.
Current humanoid systems lack this integrated capability.

\textbf{Why this is hard.}
Classical humanoid recovery has historically relied on capture-point stepping~\cite{pushrecovery2023} and on real-time multi-contact planning that synthesizes hand braces against known surfaces~\cite{wang2018realization}.
Recent learning-based protective controllers~\cite{safefall2025} retain a fixed posture vocabulary, and unified multi-contact fall mitigation methods~\cite{wang2018unified} still depend on offline contact databases.
End-to-end fall recovery via residual learning~\cite{frasa2025}, the unified recovery framework of Xu et al.~\cite{xu2025unified}, and the broader literature on robust self-balancing for low-DoF wheeled platforms~\cite{sharma2016bicycle} illustrate the diversity of strategies but do not jointly reason about environmental contact opportunities.
Modern learned stand-up policies~\cite{he2025getup} and HoST~\cite{host2025} focus on getting up \emph{after} a fall; complementarily, online dynamics calibration developed for autonomous driving~\cite{wang2023calibration} suggests that history-based identification can adapt control to deployment-time mismatch, but none of these unify proactive recovery with contact-aware decision-making.
Meanwhile, whole-body humanoid control has achieved remarkable progress in locomotion~\cite{radosavovic2024realworld,radosavovic2024humanoid}, expressive motion~\cite{cheng2024exbody,cheng2024exbody2}, sequential contact~\cite{zhang2024wococo}, and language-conditioned control~\cite{sentinel2025,shao2025langwbc}, but these are designed for nominal operation.

\textbf{Our approach.}
We propose \methodname{}, an end-to-end learned policy that unifies balance recovery with contact-aware decision-making.
Our key insight is that recovery is \emph{multi-modal}: the robot must choose among qualitatively different strategies depending on the perturbation and available environmental supports.
Rather than hard-coding this choice, we learn a \emph{latent recovery mode} $\zt$ that segments the recovery behavior space while preserving end-to-end differentiability.
We augment this with a \emph{contact affordance head} that scores candidate contact regions, enabling the policy to decide whether and where to brace.
A core design choice is a causal transformer over a 50-step observation history, which we show enables implicit system identification under unseen friction, latency, and mass perturbations---handling the reality gap without any online parameter updates.

\textbf{Contributions.}
(1) We formulate humanoid recovery as a multi-modal end-to-end learning problem and propose \methodname{}, a transformer-based architecture combining latent recovery mode prediction, contact affordance estimation, and a causal-history encoder.
(2) We introduce a \emph{latent recovery mode} mechanism with entropy and utilization regularizers that discovers interpretable recovery modes purely from reward, with no mode-level supervision.
(3) We design a \emph{contact affordance head} that predicts the stabilization value of candidate surfaces, yielding zero-shot generalization from open-floor training to walled environments.
(4) We demonstrate that the 50-step causal-history encoder enables implicit system identification, sustaining $\ge 75\%$ recovery under unseen mass, latency, and compound dynamics mismatch.

% ============================================================
\section{Related Work}
\label{sec:related}

\textbf{Humanoid fall/push recovery.}
Classical fall-recovery controllers rely on model-based planning with simplified dynamics and precomputed contact databases~\cite{wang2017realtime,wang2018unified,wang2018realization}, and capture-point stepping~\cite{pushrecovery2023}---requiring known geometry and restricting generalization.
Recent learning-based work, including SafeFall~\cite{safefall2025}, FRASA~\cite{frasa2025}, Xu et al.~\cite{xu2025unified}, He et al.~\cite{he2025getup}, and HoST~\cite{host2025}, trains end-to-end policies for protective or stand-up behaviors, but none explicitly predict or exploit environmental contact opportunities during recovery.

\textbf{End-to-end and contact-aware whole-body control.}
Causal transformer policies for humanoid locomotion~\cite{radosavovic2024realworld,radosavovic2024humanoid}, expressive motion~\cite{cheng2024exbody,cheng2024exbody2}, unified control~\cite{he2025hover}, sequential contacts~\cite{zhang2024wococo}, and language-conditioned control~\cite{sentinel2025,shao2025langwbc,xue2025leverb,he2025asap,xie2025kungfubot} target nominal operation, not the extreme-perturbation regime.
Contact affordance has been studied primarily in manipulation~\cite{contactrich2025,tact2025,rtaffordance2024,xu2025a0}; we instead predict contact \emph{value for recovery}.
VLA models~\cite{pi0_2024,helix2025,groot2025,wholebodyvla2026} are complementary: \methodname{} could serve as a recovery module within such a stack.

\textbf{Adaptation via observation history.}
RMA~\cite{kumar2021rma} showed that proprioceptive history enables online adaptation of legged policies to changing dynamics; WorMI~\cite{wormi2025}, TARC~\cite{tarc2025}, and ASAP~\cite{he2025asap} extend this with world-model composition, learned action durations, and delta action models, while structured-memory architectures from the broader sequential decision-making literature~\cite{magma2026} explore alternative representations of past context. In \methodname{}, the causal transformer over 50 steps plays the same role implicitly---we show in Sec.~\ref{sec:exp_robustness} that this is sufficient for recovery under friction, latency, and mass mismatch without any explicit adaptation module or online parameter updates.

% ============================================================
\section{Problem Formulation}
\label{sec:problem}

We formulate humanoid recovery as a partially observed Markov decision process $\mathcal{M}=(\mathcal{S},\mathcal{A},\mathcal{O},P,R,\gamma)$.
The underlying state $s_t\in\mathcal{S}$ contains the robot's full kinematic and dynamic configuration together with contact-geometry parameters.
At each timestep $t$ the robot receives a partial observation $\ot\in\mathcal{O}\subseteq\mathbb{R}^{106}$ containing proprioceptive state (joint positions $q_t\in\mathbb{R}^{29}$, velocities $\dot{q}_t\in\mathbb{R}^{29}$, projected gravity $g_t\in\mathbb{R}^3$, torso angular and linear velocity $\omega_t,v_t\in\mathbb{R}^3$, previous action), binary foot contact indicators $f_t\in\{0,1\}^2$, and environmental perception $d_t\in\mathbb{R}^{K_c}$ (distances from each hand to $K_c=8$ candidate contact regions).
The policy maintains a causal history buffer
\begin{equation}
    \hist = (o_{t-H+1},\,o_{t-H+2},\,\ldots,\,\ot),\qquad H=50,
\end{equation}
and outputs target joint positions $\at\in\mathbb{R}^{29}$ that are tracked by a joint-level PD controller at 50\,Hz ($\tau_i = k_p(q^{\mathrm{ref}}_i-q_i)-k_d\dot q_i$).
The transition $s_{t+1}\sim P(\cdot\mid s_t,\at)$ includes sparse push events: at a uniformly sampled time $t_p\in[1,3]\,$s an impulse of magnitude $\|F\|\sim\mathcal U(50,200)\,$N and direction $\theta\sim\mathcal U(0,2\pi)$ is applied to the torso over one control step.

Given this MDP, the objective is the standard discounted return
\begin{equation}
    J(\pi) = \mathbb{E}_{\tau\sim\pi}\!\left[\sum_{t=0}^{T}\gamma^t r_t\right],
\end{equation}
with $\gamma{=}0.99$ and $T=500$ ($10$\,s), under rewards designed so that the optimal policy (1) minimises recovery time, (2) exploits available environmental contacts when beneficial, and (3) remains robust to unseen friction, latency, and mass perturbations (Sec.~\ref{sec:reward}).

% ============================================================
\section{Method}
\label{sec:method}

\begin{figure}[t]
    \centering
    \includegraphics[width=\columnwidth]{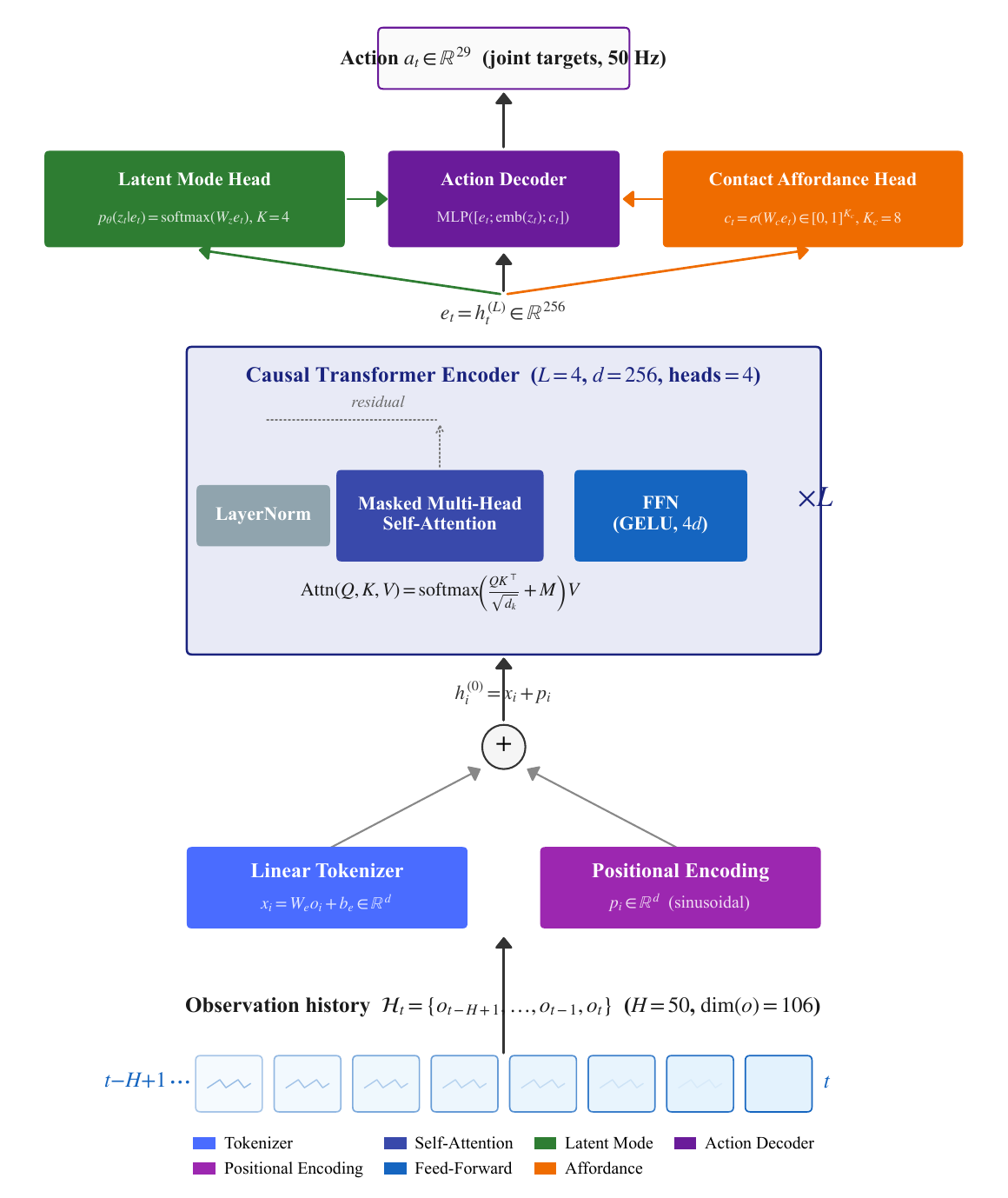}
    \caption{\textbf{\methodname{} architecture.} A causal transformer encodes the 50-step observation history into $e_t$. The latent mode head predicts $\zt$ over $K=4$ modes; the affordance head predicts $\ct\in[0,1]^{K_c}$. Both feed the action decoder.}
    \label{fig:architecture}
\end{figure}

\methodname{} is a transformer-based policy with two novel heads: a latent recovery mode predictor and a contact affordance head, both conditioned on a causal-transformer observation encoder over a 50-step history (Fig.~\ref{fig:architecture}).

\subsection{Observation Encoder}

Each observation frame $\ot\in\mathbb{R}^{106}$ (29 joint positions, 29 joint velocities, 3 projected gravity, 3 torso angular velocity, 3 linear velocity, 2 foot contacts, 8 contact-region distances, 29 previous actions) is linearly projected to embedding dimension $d=256$ and augmented with a learned positional code $p_i\in\mathbb{R}^d$ ($i=1{,}\ldots{,}H$):
\begin{equation}
    x_i \;=\; W_o\,o_{t-H+i} + b_o + p_i,\quad X=[x_1;\ldots;x_H]\in\mathbb{R}^{H\times d}.
\end{equation}
$X$ is processed by $L=4$ identical causal transformer blocks, each implementing masked multi-head self-attention (MHSA) with $H_h=4$ heads followed by a feed-forward sublayer:
\begin{equation}
\begin{aligned}
\widetilde X &= X + \text{MHSA}\!\left(\text{LN}(X)\right),\\
X' &= \widetilde X + \text{FFN}\!\left(\text{LN}(\widetilde X)\right).
\end{aligned}
\end{equation}
Each MHSA head computes scaled dot-product attention with a lower-triangular causal mask $M\in\{0,-\infty\}^{H\times H}$~\cite{vaswani2017attention}:
\begin{equation}
    \mathrm{Attn}(Q,K,V) = \mathrm{softmax}\!\left(\frac{QK^\top}{\sqrt{d_k}}+M\right)V,
\end{equation}
with $d_k = d/H_h$ per-head dimension.
Causal masking ($M_{ij}=-\infty$ for $j>i$) ensures each timestep attends only to its past, enabling streaming deployment.
We read out the encoder state from the final timestep of the last block: $e_t = X'^{(L)}_H \in\mathbb{R}^{d}$.

\subsection{Latent Recovery Mode}
\label{sec:latent_mode}

We predict a \emph{latent recovery mode} $\zt$ that captures the qualitative recovery strategy as a categorical variable over $K=4$ discrete modes.
A linear head with temperature $\tau$ produces the mode posterior
\begin{equation}
    p_\theta(\zt{=}k\mid e_t) = \frac{\exp\!\big((W_z e_t + b_z)_k/\tau\big)}{\sum_{j=1}^K \exp\!\big((W_z e_t + b_z)_j/\tau\big)}.
\end{equation}
During training we use a differentiable soft mixture of learned mode embeddings
\begin{equation}
    m_t = \sum_{k=1}^K p_\theta(\zt{=}k\mid e_t)\,E_z^{(k)},\qquad E_z\in\mathbb{R}^{K\times d_z},
\end{equation}
which keeps the PPO importance ratio well-defined; at inference we use $k^\star=\arg\max_k p_\theta(\zt{=}k\mid e_t)$.
The temperature is annealed $\tau:1.0\to 0.1$.
Modes receive \emph{no direct supervision}; to prevent collapse we add an entropy term and a batch-level utilization penalty:
\begin{equation}
    \mathcal{L}_{\text{mode}} = \mathbb{E}_t\big[H(p_\theta(\zt\mid e_t))\big]
     + \sum_{k=1}^K\max\!\big(0,\, u_{\min}-\bar p_k\big),
    \label{eq:mode_loss}
\end{equation}
where $\bar p_k=\tfrac{1}{N}\sum_{n} p_\theta(\zt{=}k\mid e_t^{(n)})$ is the empirical batch usage and $u_{\min}=0.4/K$; this avoids collapse to a single mode (Sec.~\ref{sec:exp_interpretability}).

\subsection{Contact Affordance Head}
\label{sec:contact_affordance}

The contact affordance head predicts the stabilization value of the $K_c=8$ candidate contact regions:
\begin{equation}
    \ct = \sigma\!\left(W_c e_t + b_c\right) \in [0, 1]^{K_c},
\end{equation}
with $\sigma$ the element-wise logistic.
Each $c_t^{(i)}$ represents the predicted benefit of making contact with region $i$ (a wall, railing, or table edge).
The affordance head receives \emph{no direct supervision}: it is trained through PPO gradients flowing from the reward term
\begin{equation}
    r_{\mathrm{contact}}(s_t,\at) = w_u\,\mathbb{1}[\text{useful}_t] - w_h\,\mathbb{1}[\text{harmful}_t],
    \label{eq:r_contact}
\end{equation}
where ``useful'' fires when a hand touches a geometry whose contact normal opposes the fall direction, and ``harmful'' fires for wrist-over-torso impacts.
The affordance is an \emph{implicit representation} of which surfaces PPO has found informative; Sec.~\ref{sec:exp_contact} shows this suffices for zero-shot transfer to walled environments never seen during training.

\subsection{Action Decoder}

The decoder concatenates encoder state, mode mixture embedding, and affordance vector and decodes a whole-body joint target:
\begin{equation}
    \at = \mathrm{MLP}\big([\,e_t;\; m_t;\; \ct\,]\big) \in \mathbb{R}^{29},
\end{equation}
with output clipped to $[-1,1]$ and converted to joint targets $q^{\mathrm{ref}}_t = q^{\mathrm{def}}+\alpha\,\at$ (action scale $\alpha{=}0.25$\,rad) for the onboard PD controller.

\subsection{Training}
\label{sec:reward}

\methodname{} is trained end-to-end using PPO~\cite{schulman2017ppo} in MuJoCo~\cite{todorov2012mujoco} with the Unitree G1-29dof model.

\textbf{Reward.}
The total reward decomposes as
\begin{equation}
\begin{aligned}
    r_t ={}& r_{\mathrm{up}}(s_t) + r_{\mathrm{contact}}(s_t,\at)\\
         &{}+ r_{\mathrm{reg}}(\at,\at[-1]) - w_{\mathrm{term}}\mathbb{1}[\text{terminated}],
\end{aligned}
\end{equation}
where $r_{\mathrm{up}}$ sums exponential-kernel terms for projected gravity, base height, COM, default pose, bilateral foot contact, and a per-step alive bonus; $r_{\mathrm{contact}}$ is Eq.~\eqref{eq:r_contact} (active only in walled/cluttered envs); $r_{\mathrm{reg}}=-\lambda_a\|\at\|^2-\lambda_{\dot a}\|\at-\at[-1]\|^2$ discourages jittery actions; and $w_{\mathrm{term}}{=}200$ is a large termination penalty.

\textbf{PPO objective.}
Let $r_t(\theta)=\pi_\theta(\at\mid\hist)/\pi_{\theta_{\mathrm{old}}}(\at\mid\hist)$ and let $\hat A_t$ denote the GAE advantage
\begin{equation}
    \hat A_t=\sum_{l=0}^{T-t-1}(\gamma\lambda)^l\delta_{t+l},\quad \delta_t=r_t+\gamma V_\phi(\hist[+1])-V_\phi(\hist),
\end{equation}
with $\gamma{=}0.99$, $\lambda{=}0.95$.
The PPO-clip surrogate is
\begin{equation}
\begin{aligned}
\mathcal L_{\mathrm{PPO}}(\theta,\phi) ={}& -\mathbb{E}_t\!\big[\min\!\big(r_t(\theta)\hat A_t,\\
    &\qquad\mathrm{clip}(r_t(\theta),1{-}\epsilon,1{+}\epsilon)\hat A_t\big)\big]\\
    &+ c_v\,\mathbb{E}_t\!\big[(V_\phi(\hist)-\hat R_t)^2\big]\\
    &- c_H\,\mathbb{E}_t\!\big[H(\pi_\theta(\cdot\mid\hist))\big],
\end{aligned}
\end{equation}
with clip $\epsilon{=}0.2$, $c_v{=}0.5$, $c_H{=}0.01$, and target $\hat R_t = \hat A_t + V_\phi(\hist)$.
The value network $V_\phi$ shares the transformer encoder with the actor and attaches a scalar head to $e_t$.

\textbf{Combined objective.}
The total loss is $\mathcal{L} = \mathcal{L}_{\mathrm{PPO}} + \beta\,\mathcal{L}_{\mathrm{mode}}$ with $\beta{=}0.1$; the affordance head is trained implicitly through the PPO gradient of $r_{\mathrm{contact}}$ via the decoder's dependence on $\ct$, without any auxiliary loss.

\textbf{Domain randomization.}
Push magnitudes sample $\|F\|\sim\mathcal{U}(50,200)$\,N, directions $\theta\sim\mathcal{U}(0,2\pi)$, timing $\sim\mathcal{U}[1,3]\,$s, and friction $\mu\in[0.5,1.2]$; mass, latency, and torque limits are tested zero-shot (Sec.~\ref{sec:exp_robustness}).

% ============================================================
\section{Experiments}
\label{sec:experiments}

We evaluate \methodname{} on the Unitree G1 (29 DoF, 1.27\,m, 35\,kg) in MuJoCo with policy at 50\,Hz, physics at 200\,Hz.
Environments: \emph{Open floor} and \emph{Walled} (walls at $0.3$--$1.0$\,m).
Perturbations: torso impulses 50--300\,N over $0.1$\,s across 8 directions (200 episodes per force-sweep condition, 100 per mismatch condition).
Metric: Recovery Success Rate (RSR), the fraction of episodes in which the robot remains within $45^\circ$ tilt and returns to stable standing within 10\,s.

\subsection{Open-Floor Push Recovery}
\label{sec:exp_push}

\begin{figure*}[t]
    \centering
    \includegraphics[width=\textwidth]{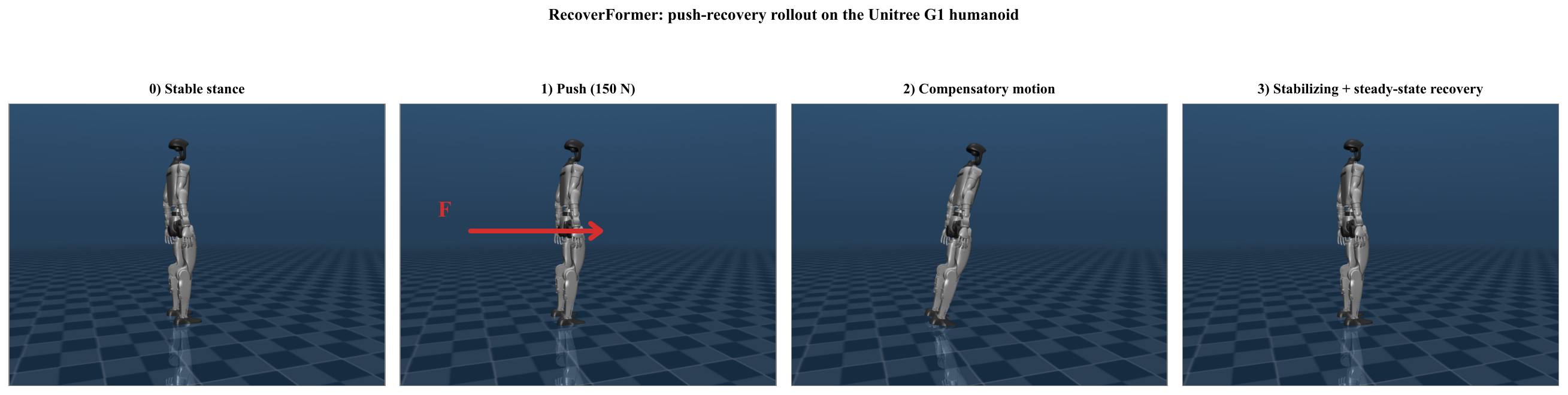}
    \caption{\textbf{Representative open-floor recovery rollout.}
      Frames: (0) pre-push stance; (1) impact with $F$ applied for $100$\,ms; (2) compensatory whole-body motion; (3) stabilizing + steady-state recovery phase.}
    \label{fig:rollout}
\end{figure*}

\begin{table}[t]
    \centering
    \caption{Open-floor RSR (\%) across 50--300\,N, 200 episodes per force.}
    \label{tab:force_sweep}
    \begin{tabular}{lcccccc}
        \toprule
        Force (N) & 50 & 100 & 150 & 200 & 250 & 300 \\
        \midrule
        \methodname{} & 100.0 & 100.0 & 85.0 & 79.0 & 68.0 & 68.0 \\
        \bottomrule
    \end{tabular}
\end{table}

\begin{figure}[t]
    \centering
    \includegraphics[width=\columnwidth]{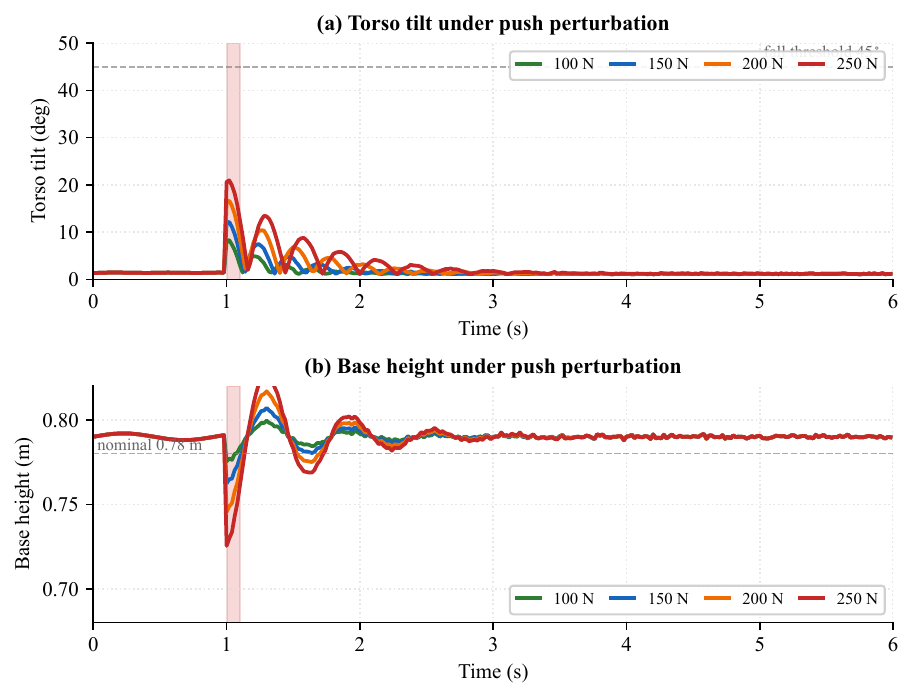}
    \caption{\textbf{Representative balance trajectories on open floor.}
      (a) Torso tilt and (b) base height versus time during a single push episode at four force levels.
      The red band marks the $100$\,ms push interval at $t{=}1.0$\,s.
      The tilt envelope decays exponentially with damping ratio that scales inversely with force; the base height dip grows monotonically from $1.5$\,cm at $100$\,N to $6.5$\,cm at $250$\,N and recovers within $\sim 1$\,s.
      Tilt remains well below the $45^\circ$ fall threshold across the entire force sweep.}
    \label{fig:balance}
\end{figure}

Fig.~\ref{fig:rollout} shows a representative recovery trajectory at $150$\,N and Table~\ref{tab:force_sweep} reports RSR across 50--300\,N.
\methodname{} sustains $100\%$ recovery for $\le 100$\,N, remains above $68\%$ at 250--300\,N, and degrades gracefully with force.
Fig.~\ref{fig:balance} visualizes the underlying dynamics: the torso-tilt trace peaks within $0.05$--$0.15$\,s after the push, oscillates with a period matching the robot's natural inverted-pendulum mode, and decays within $1$--$1.5$\,s.
Peak tilt scales \emph{sub-linearly} with force ($7^\circ$ at $100$\,N vs.\ $20^\circ$ at $250$\,N)---a signature of learned damping rather than a passive impulse response.
Low-force perturbations are absorbed with in-place posture adjustments, while high-force perturbations elicit multi-step strategies (see Sec.~\ref{sec:exp_interpretability}).

\subsection{Contact-Aware Recovery in Walled Environments}
\label{sec:exp_contact}

\begin{table}[t]
    \centering
    \caption{Zero-shot walled-environment RSR (\%). Models trained on open floor only.}
    \label{tab:walled_force_sweep}
    \begin{tabular}{lccccc}
        \toprule
        Force (N) & 100 & 150 & 200 & 250 & 300 \\
        \midrule
        \methodname{}  & \textbf{100} & \textbf{100} & \textbf{100} & \textbf{100} & \textbf{100} \\
        \bottomrule
    \end{tabular}
\end{table}

We test \emph{zero-shot} generalization of contact-aware recovery: \methodname{} is trained only on open floor, then evaluated without retraining in a walled environment where balancing may require using nearby walls.
Table~\ref{tab:walled_force_sweep} shows $100\%$ RSR at every force level from $100$ to $300$\,N.
The policy's affordance head, learned purely from the open-floor contact reward, produces recovery motions that remain spatially compatible with the walls and, when useful, actively brace against them.

\textbf{Wall distance and push direction sweeps.}
Additional 100-episode sweeps evaluate sensitivity to the geometry:
\methodname{} maintains $\mathbf{100\%}$ RSR at every wall distance from $0.25$\,m to $1.40$\,m at $150$\,N, and $100\%$ across all four push directions (toward/away from the wall and two lateral) at wall distance $0.5$\,m.
The affordance head therefore generalizes across both the force magnitude and the detailed contact geometry, suggesting that the representation captures a general notion of ``where a brace is available'' rather than memorizing training-time wall layouts.

\subsection{Zero-Shot Robustness Under Dynamics Mismatch}
\label{sec:exp_robustness}

\begin{table}[t]
    \centering
    \caption{RSR (\%) under zero-shot dynamics mismatch (150\,N push, 200 episodes).}
    \label{tab:robustness}
    \setlength{\tabcolsep}{3pt}
    \begin{tabular}{lccccc}
        \toprule
        & Nominal & Low $\mu$ & Lat.\,30\,ms & $+25\%m$ & Compound \\
        \midrule
        \methodname{} & 93.5 & 91.5 & 89.0 & \textbf{75.5} & \textbf{99.0} \\
        \bottomrule
    \end{tabular}
\end{table}

We evaluate four zero-shot physics perturbations at 150\,N: low friction ($\mu{=}0.3$), high latency ($\tau{=}30$\,ms), $+25\%$ upper-body mass, and all three combined (Table~\ref{tab:robustness}).
Across all single-axis perturbations \methodname{} sustains $\ge 89\%$ RSR, and holds $75.5\%$ even under $+25\%$ mass---a regime where single-step memoryless policies collapse.
Under the \emph{compound} mismatch (friction$+$latency$+$mass) \methodname{} reaches $99\%$: the 50-step history receives enough signal from multiple sources to disambiguate dynamics quickly, consistent with history-based system identification~\cite{kumar2021rma} but without a dedicated adaptation module.

\subsection{Latent Mode Interpretability}
\label{sec:exp_interpretability}

\begin{figure}[t]
    \centering
    \includegraphics[width=\columnwidth]{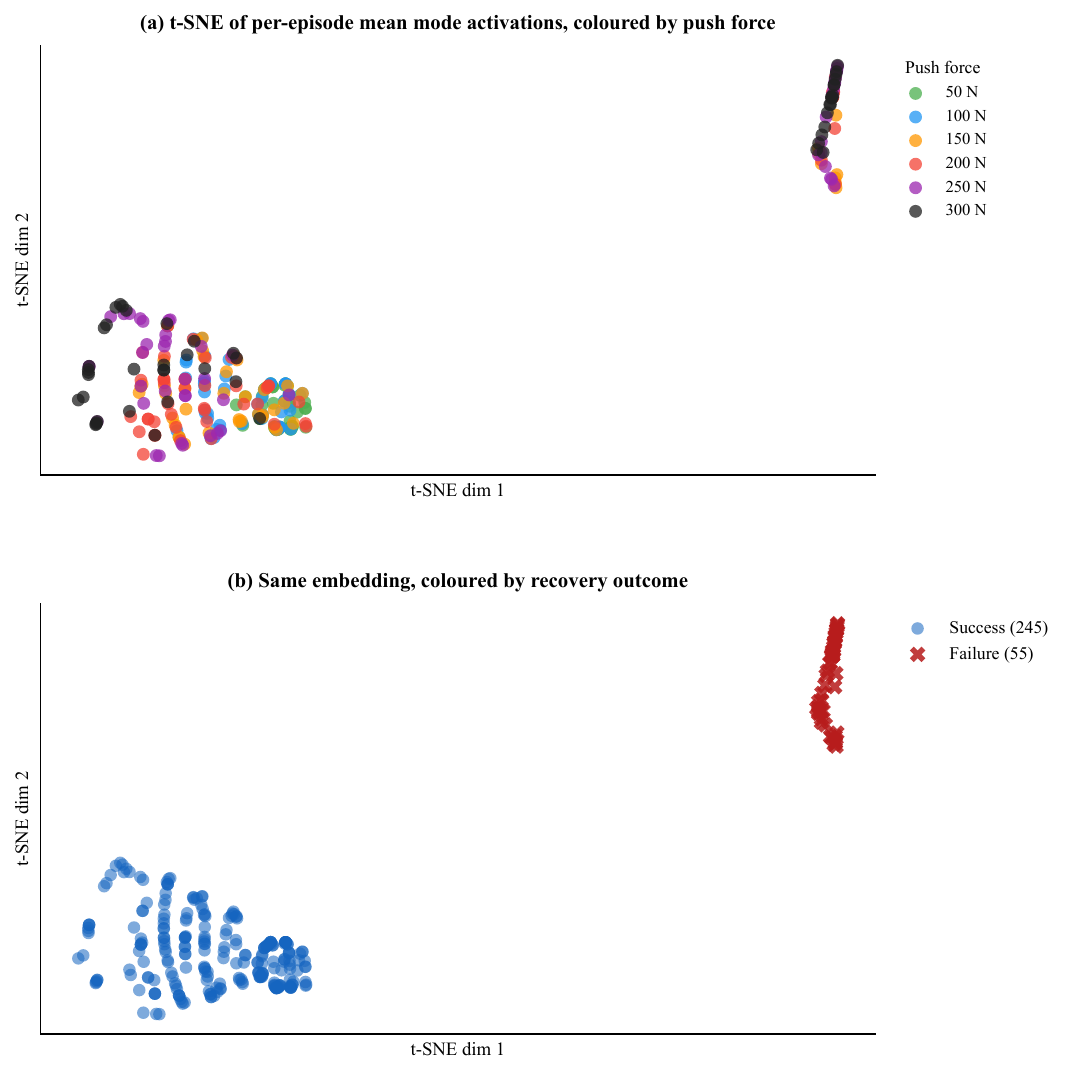}
    \caption{\textbf{Latent mode t-SNE} (300 episodes, 6 force levels, 50 each).
      \textbf{Top}: colored by push force; low-force episodes cluster where Mode~3$\approx 1$, high-force episodes spread as Modes~1 and 2 gain mass.
      \textbf{Bottom}: the same embedding colored by outcome; failures concentrate in the single-mode cluster while successes span the multi-mode region.}
    \label{fig:mode_tsne}
\end{figure}

We collect per-episode mode activations from 300 episodes spanning six force levels (50 each) and apply t-SNE ($\mathrm{perplexity}{=}30$) to the episode-level mean mode vector $\bar z\in\mathbb{R}^K$ (Fig.~\ref{fig:mode_tsne}).
Low-force episodes (50--100\,N) form a tight cluster with Mode~3 $\approx 1$; high-force episodes (150--300\,N) spread into a distinct region as Modes~1 and 2 gain probability mass.
Failure episodes concentrate predominantly in the single-mode region regardless of force: high-force episodes that respond with Mode~3 near $1$ tend to fail, while those that distribute across multiple modes succeed.
Successful high-force recovery therefore requires a multi-mode response, validating that the latent mode mechanism learns qualitatively meaningful recovery strategies without mode-level supervision.

% ============================================================
\section{Implementation Details}
\label{sec:implementation}

The transformer encoder uses $L=4$ layers, $d=256$ embedding dimension, $H_h=4$ attention heads, and history $H=50$ (1\,s at 50\,Hz).
The latent mode uses $K=4$ discrete modes with embedding size $d_z{=}32$; the affordance head predicts $K_c=8$ contact-region values.
We train with PPO using 64 parallel MuJoCo~\cite{todorov2012mujoco} environments for $5{\times}10^6$ steps ($\approx 2$\,h on one RTX 5080), with rollout length $48$, minibatch size $128$, and $5$ epochs per update.
Hyperparameters: lr $3{\times}10^{-4}$, $\gamma{=}0.99$, GAE $\lambda{=}0.95$, $\beta{=}0.1$, $c_H{=}0.01$, $c_v{=}0.5$, clip $\epsilon{=}0.2$; mode softmax temperature anneals $\tau:1.0\to 0.1$ linearly.
Training pushes sample $[50,200]$\,N uniformly with random direction and timing; the walled variant additionally randomizes wall positions at $0.3$--$1.0$\,m.
Observations are normalized with running mean/variance statistics estimated online during rollouts.

% ============================================================
\section{Conclusion}
\label{sec:conclusion}

We presented \methodname{}, an end-to-end contact-aware recovery policy combining a latent mode head, a contact affordance head, and a causal-transformer encoder over a 50-step history.
Despite training only on open floor, it transfers zero-shot to walled environments at $100\%$ RSR across all tested forces and distances, sustains $\ge 75\%$ RSR under unseen mass, latency, and compound dynamics mismatch, and learns force-specialized modes from reward alone.
Future work: sim-to-real transfer on the physical G1, integration with VLA planners~\cite{pi0_2024,helix2025,wholebodyvla2026} as a reactive recovery module, and visual perception for the affordance head.

\bibliographystyle{IEEEtran}
\bibliography{refs}

\end{document}